\documentclass[runningheads]{llncs}
\usepackage{amsmath}
\usepackage{amssymb}
\usepackage[T1]{fontenc}
\usepackage{graphicx}
\usepackage{marvosym}
\usepackage{multirow}
\usepackage[table]{xcolor}
\usepackage[
    colorlinks=true,
    linkcolor=blue,
    citecolor=blue,
    urlcolor=blue
]{hyperref}

\begin{document}

\title{DINO-Nucleus: Multi-Level Self-Supervised Features for Nucleus Instance Segmentation}

\titlerunning{DINO-Nucleus for Nucleus Instance Segmentation}

\author{
Mohammad Sarhangzadeh\inst{1}\textsuperscript{(\Letter)}
\and
Elyar Esmaeilzadeh\inst{1}
\and
Doruk \"{O}ner\inst{1}
}

\authorrunning{M. Sarhangzadeh et al.}

\institute{
NeuraVision Lab, Department of Computer Engineering, Bilkent University, Ankara, Turkey\\
\email{\{m.sarhangzadeh, elyar, doruk.oner\}@bilkent.edu.tr}
}

\maketitle

\begin{abstract}
Nucleus instance segmentation and counting are essential for quantitative microscopy, enabling reliable analysis of cell populations, tissue organization, and disease-related changes. However, accurately separating touching nuclei remains difficult, and many existing approaches rely on complex models or sensitive post-processing procedures.
We present DINO-Nucleus, a parameter-efficient framework that leverages multi-level self-supervised DINOv2 representations for nucleus instance segmentation and counting. The model jointly predicts nucleus regions, pixel-wise instance flows, and explicit nucleus centers, while a consistency objective encourages the predicted flows to converge toward their corresponding centers. Multi-level transformer features provide complementary spatial and semantic information, and the pretrained backbone is adapted using lightweight LoRA modules to reduce fine-tuning cost. Experimental results show that DINO-Nucleus achieves strong segmentation and counting performance with parameter-efficient backbone adaptation. These findings demonstrate that parameter-efficient adaptation of multi-level self-supervised features can provide an effective solution for nucleus instance segmentation in microscopy images.

\end{abstract}

\section{Introduction}
Microscopy imaging plays a central role in studying cellular organization and tissue structure. Among the many cellular components visible in microscopy images, nuclei are particularly informative because their number, distribution, size, and morphology provide valuable clues about cell populations and biological conditions \cite{caicedo2019nucleus}. Reliable nucleus analysis therefore requires not only detecting nucleus pixels, but also separating each nucleus as an individual instance so that accurate morphological measurements and cell counts can be obtained. Despite its importance, nucleus instance segmentation remains challenging because nuclei can vary considerably in shape, size, intensity, and appearance across images \cite{kumar2017dataset}. The problem becomes more difficult in densely populated regions, where neighboring nuclei may touch or overlap and their boundaries may be faint or unclear \cite{graham2019hover,stringer2021cellpose}. In such cases, semantic segmentation can identify the overall nucleus region while still merging several nuclei into a single object, so accurate counting depends on preserving instance-level structure rather than only distinguishing foreground from background.

Existing methods have substantially improved nucleus segmentation, yet reliable instance separation remains difficult in crowded images \cite{graham2019hover,horst2024cellvit}. Modern convolutional and transformer-based models can produce accurate foreground predictions, but this does not necessarily translate into correct instance boundaries or counts when adjacent nuclei share similar appearance and weak separation cues. As a result, many pipelines still depend on tuned post-processing or task-specific reconstruction rules, whose performance may vary with nucleus size, density, and image characteristics \cite{zheng2024tcunet,stringer2021cellpose}. Pretrained vision transformers provide transferable visual representations \cite{oquab2023dinov2}, but adapting them for dense prediction can require considerable computation, particularly when annotated microscopy data are limited \cite{stillwagon2026dinocell}. Moreover, many fine-tuned architectures use only the final transformer output, even though intermediate layers retain complementary spatial information useful for resolving small and closely packed structures \cite{archit2025musam,stillwagon2026dinocell}.

To address these limitations, we introduce \textbf{DINO-Nucleus}, a nucleus instance-segmentation and counting framework that leverages DINOv2 representations \cite{oquab2023dinov2}. Instead of relying only on the final transformer output, the method combines features from several backbone depths to preserve fine spatial cues alongside stronger semantic information. The pretrained encoder is adapted through lightweight LoRA modules \cite{hu2022lora}, reducing fine-tuning cost while retaining the benefits of large-scale visual pretraining. The instance-separation strategy is based on the idea that every pixel belonging to a nucleus should point toward a common center \cite{stringer2021cellpose,he2021cdnet}. The model therefore learns a pixel-wise flow field together with explicit nucleus-center predictions, while a consistency objective encourages the predicted flows to lead toward the corresponding center regions. During inference, pixels following the same flow pattern are grouped into instances, allowing touching nuclei to be separated without watershed-based reconstruction.

We evaluate DINO-Nucleus on the BBBC038 2018 dataset \cite{caicedo2019nucleus}. By combining multi-level DINOv2 features, parameter-efficient LoRA fine-tuning, and center-consistent flow learning, the proposed method achieves strong nucleus instance-segmentation and counting performance while updating only a small fraction of the backbone parameters. Together, these components provide a unified framework for transferable representation learning, parameter-efficient adaptation, and instance-aware geometric reconstruction.

\section{Related Work}

\subsection{Nucleus Instance Segmentation}

Early deep-learning approaches to nucleus segmentation were largely based on U-Net-style architectures that predict foreground masks and recover individual nuclei through connected components, contour processing, distance transforms, or watershed \cite{ronneberger2015unet,kumar2017dataset}. Within this family, methods such as TC-UNet++ \cite{zheng2024tcunet} improve semantic segmentation and instance separation through stronger feature extraction and watershed-based reconstruction. Proposal-based methods such as Mask R-CNN \cite{he2017mask} take a different approach by detecting candidate objects and predicting a separate mask for each proposal, although densely packed and small nuclei can remain difficult to localize reliably. Nucleus-specific methods have also introduced geometric representations that encode instance structure more directly \cite{he2021cdnet}. StarDist \cite{schmidt2018cell} represents each nucleus using a center probability and radial distances that form a star-convex polygon, while HoVer-Net predicts horizontal and vertical center-relative maps to separate touching nuclei \cite{graham2019hover}. CellViT combines pretrained transformer encoders with a HoVer-style decoder, demonstrating how stronger visual representations can be integrated with vector-based nucleus reconstruction \cite{horst2024cellvit}. Flow-based methods provide another alternative by assigning each foreground pixel a direction toward a shared convergence region. Cellpose popularized this strategy for biological instance segmentation and reconstructs instances by grouping pixels according to their predicted trajectories \cite{stringer2021cellpose}. However, the nucleus center is generally represented only implicitly through the learned flow field \cite{stillwagon2026dinocell}.

\subsection{DINO and Parameter-Efficient Fine-Tuning}

Pretrained vision transformers provide transferable representations that are particularly useful when task-specific annotations are limited \cite{moutakanni2025cell,stillwagon2026dinocell,lavaee2026spatialdino}. Self-supervised models such as DINO \cite{caron2021emerging} and DINOv2 \cite{oquab2023dinov2} learn without manual class labels and capture meaningful object-level structure in their attention and feature representations. Features from different transformer depths provide complementary information: intermediate layers preserve finer spatial organization, whereas deeper layers encode more abstract semantic content \cite{ranftl2021dpt}.

Adapting these large backbones nevertheless involves a trade-off between accuracy and fine-tuning cost. Full fine-tuning requires substantial memory and computation, while freezing the encoder can limit adaptation to microscopy images. Parameter-efficient fine-tuning methods address this issue by updating only a small subset of parameters. LoRA learns low-rank updates to selected transformer projections while leaving the pretrained weights unchanged \cite{hu2022lora}.

Building on both lines of work, DINO-Nucleus combines multi-level DINOv2 representations with LoRA-based adaptation for parameter-efficient feature transfer. It further connects these representations to flow-based nucleus reconstruction by jointly predicting pixel-wise flows and explicit centers and enforcing consistency between them. This framework brings together parameter-efficient pretrained feature adaptation and center-guided instance separation. 

\begin{figure}[t]
    \centering
    \includegraphics[width=\linewidth, height=6.0cm, keepaspectratio=false]{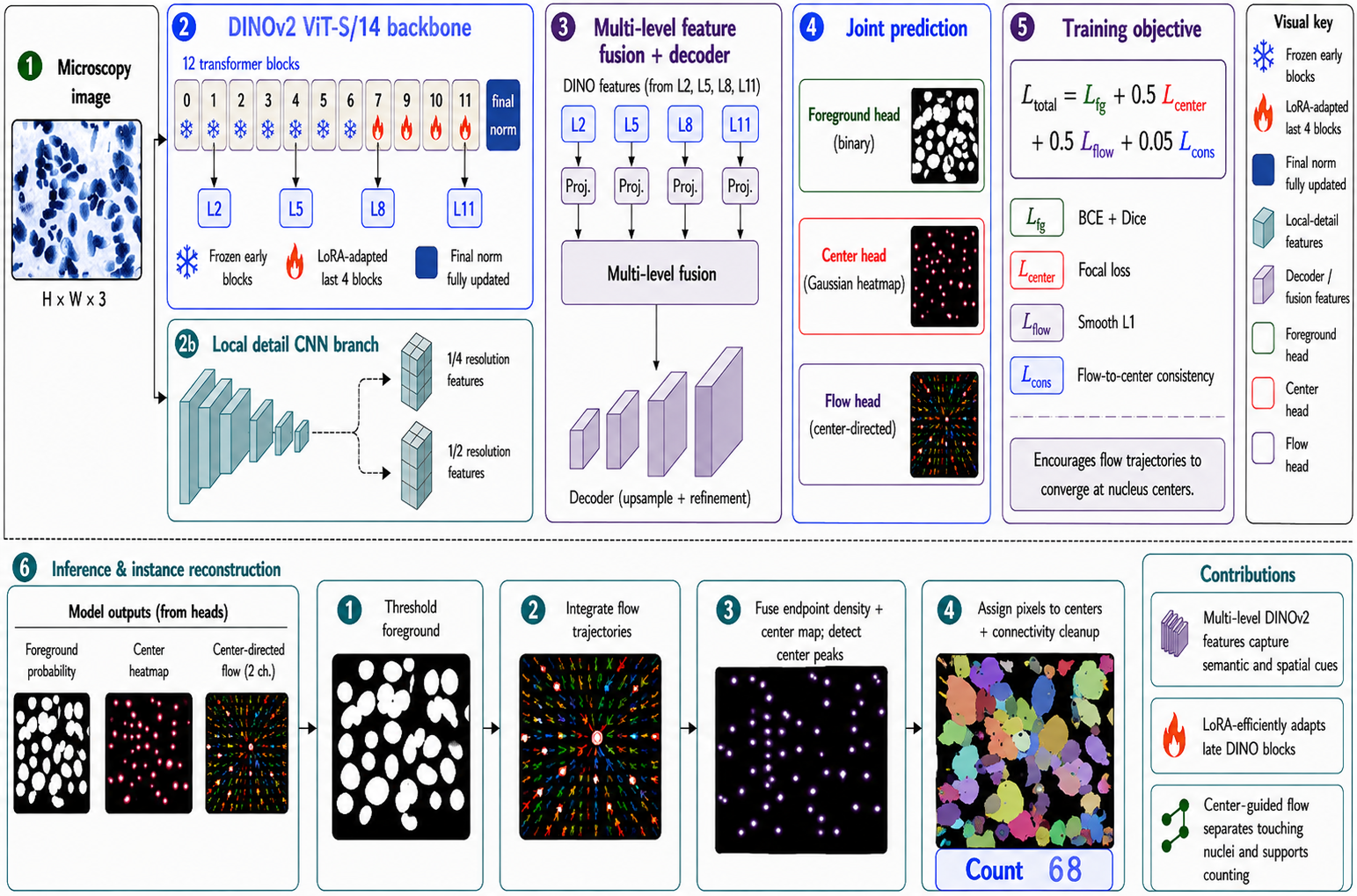}
    \caption{Overview of DINO-Nucleus, showing multi-level DINOv2 feature fusion, joint foreground, center, flow prediction, and center-guided instance reconstruction.}
    \label{fig:method_overview}
\end{figure}
\section{Method}

Given an RGB microscopy image $\mathbf{X}\in\mathbb{R}^{3\times H\times W}$, the objective is to predict an instance map $\hat{\mathbf{M}}\in\{0,1,\ldots,\hat{N}\}^{H\times W}$, where zero denotes background and each positive label identifies one nucleus. The final nucleus count $\hat{N}$ is obtained from the number of unique non-background labels in $\hat{\mathbf{M}}$.

As shown in Fig.~\ref{fig:method_overview}, DINO-Nucleus does not predict the instance map directly. Instead, the network produces a foreground probability map $\hat{\mathbf{Y}}_{\mathrm{fg}}\in[0,1]^{H\times W}$, a center heatmap $\hat{\mathbf{Y}}_{\mathrm{c}}\in[0,1]^{H\times W}$, and a two-channel center-directed flow field $\hat{\mathbf{Y}}_{\mathrm{f}}\in\mathbb{R}^{2\times H\times W}$. The foreground map identifies nucleus pixels, the center heatmap provides explicit instance locations, and the flow field directs each foreground pixel toward the center of its corresponding nucleus.

The input image is encoded by a pretrained DINOv2 backbone adapted with LoRA, while features from multiple transformer depths are retained. A lightweight CNN branch also extracts higher-resolution local information from the input. The projected transformer and CNN features are fused through a shared decoder and passed to separate foreground, center, and flow heads. Training combines foreground segmentation, center localization, and flow regression. During inference, the three outputs are jointly used to integrate pixel trajectories, detect nucleus centers, assign foreground pixels to instances, and obtain the final instance map and count.

\subsection{Encoding and Feature Fusion}

Given an input image $\mathbf{X}$, we use a pretrained DINOv2 ViT-S/14 encoder and retain features from different stages of the transformer to capture information from its early, intermediate, and final layers. Among the layer combinations considered during model development, layers $2$, $5$, $8$, and $11$ provided the best validation performance and were therefore selected. For each layer $l$, the feature map is written as $\mathbf{F}_l\in\mathbb{R}^{C\times H_p\times W_p}$, where $H_p$ and $W_p$ denote the patch-grid dimensions. Although these features share the same spatial resolution, they represent different levels of abstraction: earlier layers retain more local structure, while deeper layers provide stronger semantic context. Each feature is projected into a common channel space using a layer-specific $1\times1$ convolution, $\mathbf{P}_l=\phi_l(\mathbf{F}_l)$, before the projected features are combined into a multi-level representation.

To adapt the pretrained encoder efficiently, all early transformer blocks remain frozen, while the final four blocks are adapted using LoRA modules. The original weights of these blocks remain fixed, and only the low-rank updates are optimized. The final normalization layer is fully trainable to further align the adapted representations with the microscopy task. Since the DINOv2 features remain on the patch grid, a lightweight CNN branch also extracts higher-resolution local features $\mathbf{L}_{1/4}$ and $\mathbf{L}_{1/2}$ from the input image. The fused transformer representation is then progressively upsampled and combined with these local features through the decoder, producing a shared full-resolution representation $\mathbf{Z}$ for the foreground, center, and flow prediction heads.

\subsection{Center-Flow Learning and Instance Reconstruction}

A foreground map identifies nucleus regions but cannot reliably separate touching nuclei once they form a connected component. We therefore complement it with an explicit center heatmap and a pixel-wise center-directed flow field. Together, $\hat{\mathbf{Y}}_{\mathrm{fg}}$, $\hat{\mathbf{Y}}_{\mathrm{c}}$, and $\hat{\mathbf{Y}}_{\mathrm{f}}$ describe nucleus extent, instance locations, and pixel-to-instance associations.

The supervision targets are generated from the ground-truth instance masks. The foreground target $\mathbf{Y}_{\mathrm{fg}}$ is obtained by merging all nucleus instances into a binary mask. For each nucleus, its center $c_k$ is selected as the point with the largest Euclidean distance from the boundary, ensuring that it lies inside the instance even for irregular shapes. To provide denser and more stable center supervision, a Gaussian response is placed around each center:

\begin{equation}
\mathbf{Y}_{\mathrm{c}}(p)
=
\max_k
\exp\left(
-\frac{\|p-c_k\|_2^2}{2\sigma_k^2}
\right),
\end{equation}

where $\sigma_k$ controls the spatial extent of the response for nucleus $k$. For each pixel $p$ belonging to nucleus $k$, the flow target is defined as a normalized vector pointing toward its center:

\begin{equation}
\mathbf{Y}_{\mathrm{f}}(p)
=
\frac{c_k-p}{\|c_k-p\|_2+\epsilon}.
\end{equation}

Background flow is set to zero. This representation allows pixels from the same nucleus to converge toward a common location, while pixels from touching nuclei are directed toward different centers. The foreground branch is trained using binary cross-entropy and Dice loss, such that
$\mathcal{L}_{\mathrm{fg}}
=
\mathcal{L}_{\mathrm{BCE}}
+
\mathcal{L}_{\mathrm{Dice}}$.
The center branch uses Gaussian focal loss,
$\mathcal{L}_{\mathrm{c}}
=
\mathcal{L}_{\mathrm{GF}}
(\hat{\mathbf{Y}}_{\mathrm{c}},\mathbf{Y}_{\mathrm{c}})$.
Flow regression is restricted to foreground pixels:

\begin{equation}
\mathcal{L}_{\mathrm{f}}
=
\frac{
\sum_p \mathbf{Y}_{\mathrm{fg}}(p)
\sum_{d\in\{x,y\}}
\operatorname{SmoothL1}
\left(
\hat{\mathbf{Y}}_{\mathrm{f}}^{d}(p),
\mathbf{Y}_{\mathrm{f}}^{d}(p)
\right)
}{
\sum_p\mathbf{Y}_{\mathrm{fg}}(p)+\epsilon
},
\end{equation}

preventing the much larger background region from dominating the objective. Local flow regression alone does not ensure that the predicted trajectories terminate at valid nucleus centers. We therefore introduce endpoint consistency. Starting from a foreground pixel $p$, the predicted flow field is differentiably integrated over $T$ steps:

\begin{equation}
q_p^{0}=p,
\qquad
q_p^{t+1}
=
q_p^t+
s\,\rho(q_p^t)
\frac{
\hat{\mathbf{Y}}_{\mathrm{f}}(q_p^t)
}{
\|\hat{\mathbf{Y}}_{\mathrm{f}}(q_p^t)\|_2+\epsilon
},
\end{equation}

where $s$ is the integration step size, $\rho(q_p^t)$ denotes  foreground support sampled at current trajectory position, and $q_p^T$ is the resulting endpoint. The center reference combines the ground-truth heatmap with detached center prediction:

\begin{equation}
\mathbf{Y}_{\mathrm{ref}}
=
\eta\mathbf{Y}_{\mathrm{c}}
+
(1-\eta)\operatorname{sg}
\left(
\hat{\mathbf{Y}}_{\mathrm{c}}
\right),
\end{equation}

where $\operatorname{sg}(\cdot)$ prevents gradients from this term from updating the center branch. Endpoint consistency is then defined as

\begin{equation}
\mathcal{L}_{\mathrm{end}}
=
\frac{
\sum_p\mathbf{Y}_{\mathrm{fg}}(p)
\left[
-\log\left(
\mathbf{Y}_{\mathrm{ref}}(q_p^T)+\epsilon
\right)
\right]
}{
\sum_p\mathbf{Y}_{\mathrm{fg}}(p)+\epsilon
}.
\end{equation}

While $\mathcal{L}_{\mathrm{f}}$ supervises local flow directions, $\mathcal{L}_{\mathrm{end}}$ encourages foreground trajectories to terminate within valid nucleus-center. The complete objective is

\begin{equation}
\mathcal{L}
=
\lambda_{\mathrm{fg}}\mathcal{L}_{\mathrm{fg}}
+
\lambda_{\mathrm{c}}\mathcal{L}_{\mathrm{c}}
+
\lambda_{\mathrm{f}}\mathcal{L}_{\mathrm{f}}
+
\lambda_{\mathrm{end}}r(e)\mathcal{L}_{\mathrm{end}},
\end{equation}

where the $\lambda$ coefficients balance the loss terms and $r(e)$ gradually introduces endpoint consistency during training. Training begins with a decoder-only warm-up, after which LoRA parameters and final normalization layer are optimized.

During inference, the foreground map is thresholded to identify candidate nucleus pixels, whose trajectories are then integrated through the predicted flow field. Their endpoints are accumulated into an endpoint-density map, where high-density regions indicate likely nucleus centers. This density is combined with the predicted center heatmap, and local peaks in the fused response are selected as instance centers.

Each foreground pixel is assigned to a detected center according to its trajectory endpoint. Disconnected regions sharing the same provisional label are separated, and small isolated fragments are removed. The remaining connected regions form the final instance map $\hat{\mathbf{M}}$, from which nucleus count $\hat{N}$ is obtained.
\section{Experiments}
We describe the dataset, evaluation protocol, metrics, and implementation
settings, followed by qualitative analysis of the learned representations,
center heatmaps, and flow fields. We then compare DINO-Nucleus with
TC-UNet++, Cellpose3, StarDist, and CellViT
\cite{zheng2024tcunet,stringer2025cellpose3,schmidt2018cell,horst2024cellvit}
in terms of segmentation,
counting, and parameter efficiency. Finally, we compare it with a basic
DINOv2 watershed baseline \cite{oquab2023dinov2} to assess the contributions of center-flow consistency.

\textbf{Dataset and preprocessing.}
We use BBBC038, containing 670 microscopy images with instance-level
nucleus annotations \cite{caicedo2019nucleus}. Using a fixed seed, images are stratified by
nucleus count and mean intensity into 468 training, 67 validation, and
135 test samples. The validation set is used for model selection and
post-processing calibration, while the test set is reserved for final
evaluation. Training images are aspect-ratio resized and zero-padded to
$448\times448$, with random geometric and photometric augmentation.
Test images are processed using $448\times448$ sliding windows with
112-pixel overlap.

\textbf{Evaluation metrics.}
We evaluate foreground segmentation using Dice and IoU, and instance
segmentation using object-level F1, panoptic quality (PQ), and
aggregated Jaccard index (AJI) \cite{kirillov2019panoptic,kumar2017dataset}. Instances are matched at
$\mathrm{IoU}>0.45$. Counting is evaluated using mean absolute error
(MAE) and Pearson correlation.

\textbf{Implementation details.}
DINO-Nuclei uses a pretrained DINOv2 ViT-S/14 encoder \cite{oquab2023dinov2} and fuses features
from blocks $2$, $5$, $8$, and $11$ through a 128-channel decoder.
Rank-8 LoRA adapters with scaling 16 and dropout 0.05 are applied to
the final four blocks \cite{hu2022lora}. Foreground, center, flow, and endpoint-consistency
losses are weighted by $1.0$, $0.5$, $0.5$, and $0.05$, respectively,
with consistency enabled after five warm-up epochs. During inference, foreground pixels follow the predicted flow and are
grouped by their converged endpoints \cite{stringer2021cellpose}. Reconstruction and post-processing
parameters are calibrated on the validation set and fixed for testing. Flow reconstruction uses 48 integration steps with unit step size; the remaining reconstruction thresholds are selected on validation set and fixed for testing.
All baselines use the same splits, augmentations, epoch budget, and
evaluation protocol while retaining architecture-specific training
settings. Experiments run on an NVIDIA
A100 GPU.

\begin{figure}[!t]
    \centering
    \includegraphics[width=\linewidth]{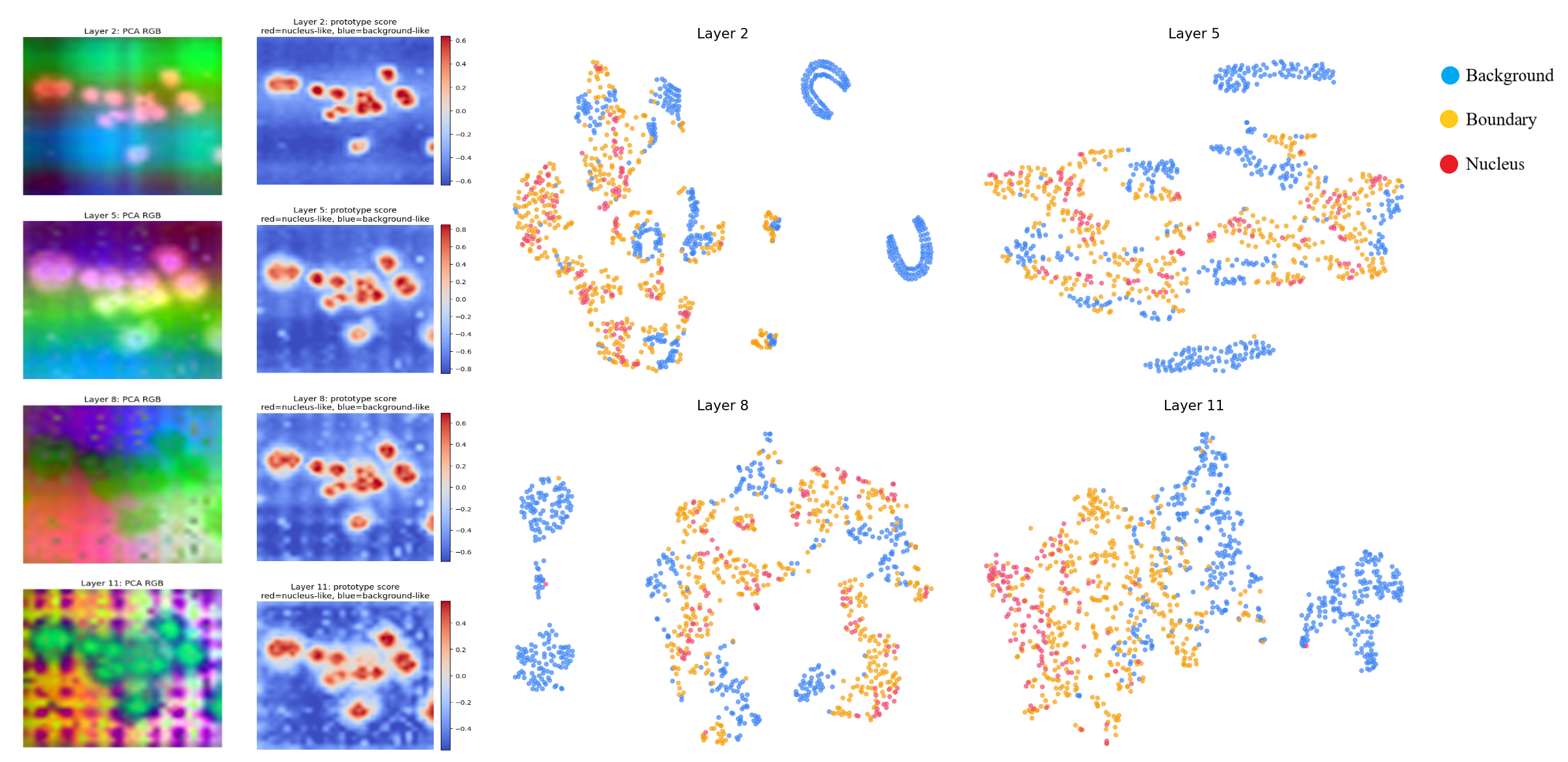}
  \caption{Pretrained DINOv2 features from layers $2$, $5$, $8$, and
$11$. PCA maps visualize spatial token organization, prototype scores
highlight nucleus (red) and background (blue) regions.
T-SNE plots show background, boundary, and nucleus-dominant patches.}
    \label{fig:dino_features}
\end{figure}

\textbf{Feature analysis.}
Figure~\ref{fig:dino_features} examines the pretrained DINOv2 features
used by DINO-Nucleus. The PCA maps show that nucleus regions form clear
patterns that differ from the surrounding background. This is also
visible in the prototype-score maps, where nucleus locations appear
nucleus-like across all four layers. The t-SNE plots provide another
view of the feature space: background patches often form separate
groups, while nucleus and boundary patches remain more closely related.
This is expected as boundary patches contain both nucleus and
background pixels. Taken together, PCA maps show that DINOv2
preserves the spatial structure of nucleus regions, while the t-SNE
plots show that nucleus-dominant patches occupy feature-space regions
that are largely separate from background patches, with boundary
patches naturally appearing between them.

\textbf{Comparison with existing methods.}
Table~\ref{tab:main_results} compares our method with existing nucleus
segmentation approaches and the DINO baseline on BBBC038. Our method achieves competitive foreground segmentation, despite
adapting only a small part of the pretrained encoder through LoRA. Together with the feature
analysis in Fig.~\ref{fig:dino_features}, this indicates that pretrained
DINOv2 features already provide strong information for locating nucleus
regions. The main advantage of our method appears in instance segmentation and
counting, where it consistently outperforms all competing methods. Compared with the DINOv2 baseline using watershed reconstruction, the
foreground results remain similar, as the center-flow formulation is
not mainly designed to improve foreground detection. In contrast, the
large improvements in instance segmentation and counting show that the
predicted centers, center-directed flows, and their consistency are
effective for separating neighboring nuclei and reconstructing
individual instances.

\begin{table}[t]
\caption{Comparison of nucleus segmentation and counting performance.}
\label{tab:main_results}
\centering
\resizebox{\textwidth}{!}{%
\begin{tabular}{l|cc|ccc|cc}
\hline
\multirow{2}{*}{\textbf{Method}}
& \multicolumn{2}{c|}{\textbf{Foreground}}
& \multicolumn{3}{c|}{\textbf{Instance}}
& \multicolumn{2}{c}{\textbf{Counting}} \\
\cline{2-8}
& \textbf{Dice $\uparrow$}
& \textbf{IoU $\uparrow$}
& \textbf{Obj.-F1 $\uparrow$}
& \textbf{PQ $\uparrow$}
& \textbf{AJI $\uparrow$}
& \textbf{MAE $\downarrow$}
& \textbf{Pearson $\uparrow$} \\
\hline
TC-UNet++
& \textbf{92.46}
& \textbf{88.51}
& 83.64
& 72.33
& 77.59
& 9.19
& 89.70 \\

Cellpose3
& 88.71
& 81.34
& 88.52
& 76.69
& 83.13
& 5.97
& 93.39 \\

StarDist
& 85.94
& 81.07
& 83.06
& 70.05
& 75.93
& 9.72
& 87.62 \\

CellViT
& 90.22
& 84.35
& \underline{90.75}
& \underline{77.40}
& \underline{85.35}
& \underline{5.63}
& \underline{94.58} \\

DINOv2
& \underline{90.98}
& \underline{86.16}
& 83.51
& 74.88
& 79.28
& 7.47
& 89.20 \\

\textbf{Ours}
& \cellcolor{green!15}90.45
& \cellcolor{green!15}85.30
& \cellcolor{green!15}\textbf{92.38}
& \cellcolor{green!15}\textbf{79.33}
& \cellcolor{green!15}\textbf{88.97}
& \cellcolor{green!15}\textbf{3.11}
& \cellcolor{green!15}\textbf{98.75} \\
\hline
\end{tabular}%
}
\end{table}

We next compare the model sizes and number of trainable parameters in
Table~\ref{tab:param_comparison}. Although DINO-Nucleus uses a
transformer encoder, most pretrained parameters remain frozen, and only
the decoder, LoRA adapters, and final normalization layer are updated.
This allows the model to achieve strong performance with limited
backbone adaptation.

\begin{table}[t]
\caption{Parameter comparison between the evaluated methods}
\label{tab:param_comparison}
\centering
\begin{tabular}{l|ccc}
\hline
\textbf{Method}
& \textbf{Total (M) $\downarrow$}
& \textbf{Trainable (M) $\downarrow$}
& \textbf{Trainable (\%) $\downarrow$} \\
\hline
TC-UNet++ & 9.22 & 9.22 & 100.00 \\
Cellpose3 & \underline{6.60} & 6.60 & 100.00 \\
StarDist  & \textbf{1.99} & \textbf{1.99} & 100.00 \\
CellViT   & 39.95 & 39.95 & 100.00 \\
DINOv2    & 24.28 & 9.32 & \underline{38.40} \\
\textbf{Ours}
& \cellcolor{green!15}24.35
& \cellcolor{green!15}\underline{4.07}
& \cellcolor{green!15}\textbf{16.72} \\
\hline
\end{tabular}
\end{table}

\textbf{Limitations.}
A limitation of the current formulation is its reliance on a single
convergence center per nucleus, which may be less suitable for highly
elongated or irregular shapes. In addition, evaluation is limited to
BBBC038, and the validation-calibrated reconstruction parameters may
require adjustment for datasets with substantially different imaging
characteristics.
\section{Conclusion}
In this paper, we introduced DINO-Nucleus, a parameter-efficient framework for nucleus instance segmentation and counting. By leveraging multi-level DINOv2 features adapted via lightweight LoRA modules, the model effectively combines fine spatial details with robust semantic context. Our approach jointly predicts foreground regions, explicit nucleus centers, and pixel-wise flows, utilizing a consistency objective that directs pixel trajectories toward valid centers. This center-guided formulation accurately separates touching nuclei without relying on sensitive watershed post-processing. Experiments on the BBBC038 dataset demonstrate that DINO-Nucleus significantly outperforms existing methods in both instance segmentation and counting accuracy. Ultimately, this highlights the strong potential of combining efficiently adapted self-supervised representations with geometric learning for reliable quantitative microscopy.

\bibliographystyle{splncs04}
\bibliography{main}

\end{document}